\documentclass{article}

\usepackage{microtype}
\usepackage{graphicx}
\usepackage{subfigure}
\usepackage{booktabs} 

\usepackage{amsmath}
\usepackage{amsfonts}
\usepackage{color}
\usepackage{blindtext}
\usepackage{soul}

\usepackage{hyperref}

\usepackage[accepted]{icml2018}
\usepackage{cleveref}

\icmltitlerunning{Sequential sampling of Gaussian latent variable models}

\newcommand{\Ordo}{\mathcal{O}}
\newcommand{\bsigma}{\boldsymbol{\sigma}}
\newcommand{\f}{\mathbf{f}}
\renewcommand{\c}{\mathbf{c}}
\newcommand{\y}{\mathbf{y}}
\newcommand{\x}{\mathbf{x}}
\newcommand{\m}{\mathbf{m}}
\newcommand{\z}{\mathbf{z}}
\newcommand{\K}{\mathbf{K}}
\newcommand{\bL}{\mathbf{L}}
\newcommand{\bnu}{\boldsymbol{\nu}}
\newcommand{\GP}{\text{GP}}

\newcommand{\calS}{\mathcal{S}}
\newcommand{\KL}{\textrm{KL}}
\newcommand{\bbE}{\mathbb{E}}

\newcommand{\IhatSGP}{\hat{I}^{\varphi_t}_{\text{SGP}}}
\newcommand{\IhatMC}{\hat{I}^{\varphi_t}_{\text{MCMC}}}
\newcommand{\I}{I^{\varphi_t}}
\newcommand{\Var}{\text{Var}}
\newcommand{\bias}{b_{\varphi_t}}
\newcommand{\mse}{\text{MSE}}

\begin{document}

\twocolumn[
\icmltitle{Sequential sampling of Gaussian process latent variable models}



\icmlsetsymbol{equal}{*}

\begin{icmlauthorlist}
\icmlauthor{Martin Tegn\'{e}r}{to}
\icmlauthor{Benjamin Bloem-Reddy}{goo}
\icmlauthor{Stephen Roberts}{to}
\end{icmlauthorlist}

\icmlaffiliation{to}{Department of Engineering Science, University of Oxford, Oxford, United Kingdom}
\icmlaffiliation{goo}{Department of Statistics, University of Oxford, Oxford, United Kingdom}

\icmlcorrespondingauthor{Martin Tegn\'{e}r}{martin.tegner@eng.ox.ac.uk}

\icmlkeywords{Latent Variable Models, Gaussian Processes, Expectation Propagation, Elliptical Slice Sampling}

\vskip 0.3in
]



\printAffiliationsAndNotice{}  

\begin{abstract}
We consider the problem of inferring a latent function in a probabilistic model of data. When dependencies of the latent function are specified by a Gaussian process and the data likelihood is complex, Bayesian inference often involves Markov chain Monte Carlo sampling, which has limited applicability to large data sets, due to computational complexity. For problems exhibiting a sequential structure, we adapt these techniques and propose an approximation that enables sequential sampling of both latent variables and associated parameters. We demonstrate strong performance in growing-data settings that would otherwise be infeasible with naive, non-sequential sampling.
\end{abstract}

\section{Introduction}

Gaussian processes ($\GP$s) are used extensively by the machine learning community as a flexible framework for non-parametric modelling \cite{rasmussen2006gaussian}. They offer a probabilistic approach to infer and predict dependencies in data.  Here we concentrate on their use 
 in latent variable models, where an unobservable function is generative of data through a possibly complex and non-linear likelihood. 
 
These models give  rise to inversion problems. At a high level, one can think of the latent quantity as a non-observable input function to a physical system, from which one has noisy measurements of the output. The measurements relate to each other in a way  prescribed by the latent function and the system, and the problem is to ``inversely'' find a function that best explains  observed data. 
 
In a probabilistic framework, the aim is to infer a distribution over the latent function, and not just a single ``best'' point-estimate. This acknowledges a notion of uncertainty attached to the estimate, which stems from  availability and noisiness of data, and the interplay of the latent function with the output-generating system itself. The $\GP$ acts as a nonparametric model  which  encodes prior beliefs and domain knowledge about the latent function. The system's input-to-output mapping is formalised in a likelihood model, which also specifies a noise distribution. Together with a prior over the model's hyperparameters, the $\GP$ and likelihood of observed data then induces a posterior distribution over the latent function. These type of models are considered within system biology \cite{barenco2006ranked}, geostatistics \cite{rue2009approximate}, and robot kinematics \cite{williams2009multi}, to mention just but a few examples.

With the exception of Gaussian likelihoods with known hyperparameters, inference on the posterior distribution of GP latent variable models is generally intractable. For this purpose, we build on Markov chain Monte Carlo (MCMC) methods, which approach inference with samples from the posterior. In particular, elliptical slice sampling (ESS) \cite{murray2010elliptical} provides a versatile  method demonstrated to be fast and efficient for a range of Gaussian latent variable models. 
ESS updates latent variables for a known (or fixed)  covariance; the related and surrogate data slice sampling (SDSS) \cite{murray2010slice} updates  hyperparameters. In cooperation they provide a MCMC  strategy for full Bayesian inference.

However, these methods are limited by their scaling with data size, $N$. Exploitable structure 
and approximations of the likelihood aside (e.g. \cite{saatcci2012scalable} and \cite{quinonero2005unifying}),
the computational cost  of decompositions required for covariance parameter updates is $\Ordo (N^3)$, which 
becomes problematic already when $N$ approaches a few thousand. 

To the best of our knowledge, there is not much previous work that addresses the issue of  computational complexity  within MCMC samplers for inference in latent $\GP$ models with sequential structure. In this paper, we provide a  strategy that aims to solve this issue  when the data generating process has a sequential structure. We look at problems of data size $TN$, where $N$ is the observation size at each of {$T$} steps. 
We propose a simple algorithm that makes sequential approximations of the posterior with expectation propagation \cite{minka2001expectation} and use this as target distribution in a MCMC sampler. This cuts the cost from $\Ordo(T^3N^3)$ for sampling from the full model to $\Ordo(T\tau^3N^3)$, for a tunable or user-specified constant $\tau$. Within the approximate sampling algorithm, we exploit the strengths of ESS and SDSS to obtain a practical and fast MCMC sampler. We demonstrate the benefits with empirical experiments  based on two cases: one standard benchmark and one  complex likelihood model.

\section{Sequential latent Gaussian process models}\label{sec:lgpm}

We consider Bayesian inference for probabilistic models of observable data $\y\in\mathbb{R}^n$. 
We assume  a given likelihood  which represents the data-generating process, $\y\sim p(\y|\f,\alpha)$, where a latent variable $\f\in\mathbb{R}^{N}$ is the main object of interest for inference. {As an example, one can have in mind a system  where $\f$ is an unobservable input and $\y$ noisy measurements of the output} $ \y = \Phi(\f) +\boldsymbol{\epsilon}$.
Parameters associated with the likelihood---{with the system $\Phi$ and noise distribution $p(\boldsymbol{\epsilon})$}---are collected in $\alpha$. 
 
We assume the latent variable  has a functional prior  $f(x)\sim\mathcal{GP}(0,k_{\kappa}(x,x'))$ with input space $x\in\mathbb{R}^D$. 
Let $\mathcal{N}(\textbf{z}|\m,\K)$ denote a Gaussian density with mean $\m$ and covariance matrix $\K$.  
By definition, the $\GP$ induces a prior on the latent variable $\f \sim p(\f|\kappa) = \mathcal{N}(\f|\mathbf{0},\K_{\kappa})$.
$\K_{\kappa}$ is the Gram matrix constructed by the covariance kernel with parameters $\kappa$: $[\K_{\kappa}]_{i,j} = k_{\kappa}(x_i,x_j)$ 
for all pairs $x_i,x_j$ of the input set $\x\in\mathbb{R}^{N\times D}$ corresponding to $\f$.
A  mean vector $\m_m$ can be absorbed into the likelihood function $p(\y|\f,\alpha) \equiv p(\y|\f+\m_m,\alpha)$, where  parameters $m$ are included in $\alpha$ for convenience. 

\textbf{Sequential data model}\, Consider now the data to be a sequence of observations $\y = (\y_t)_{t=1}^T$ with corresponding $\f=(\f_t)_{t=1}^T$, where the $\y_t\in\mathbb{R}^{n}$ are assumed conditionally independent given  $\f_t\in\mathbb{R}^{N}$. The likelihood then factorises over $t$ (but not necessarily over the {elements} of $\y_t$): $p(\y_{1:T}|\f_{1:T},\alpha) = \prod_{t=1}^T p(\y_t|\f_t,\alpha)$ 
where we use a  shorthand notation $\y_{1:T} \equiv (\y_t)_{t=1}^T$. 

In this setting, the temporal dependence in  data is induced by the dependency structure of the latent $\GP$, modelled by augmenting the input space with $t\in\mathbb{N}$. We assume a covariance kernel that is separable and isotropic in $t$
\begin{equation}\label{eqn5}
k_{\kappa}(t,x;t',x') = k^{(t)}_{\kappa}(|t-t'|)\, k^{(x)}_{\kappa}(x,x').
\end{equation}
Thus, the covariance matrix $\K_{\kappa}$ of $\f_{1:T}$ is the Gram matrix over all $(t,x)$-inputs in $\x_{1:T}\equiv(\x_t)_{t=1}^{T}$, where $\x_t$ denotes the set of spatial inputs belonging to $\f_t$.

\textbf{Kernel selection}\, 
The isotropic form \eqref{eqn5} is not  strictly necessary, although we consider it a natural assumption for sequential/temporal data. Alternatively, one may consider non-stationary covariances, such as periodic kernels. Incorporating non-stationary covariance in the present work would require more sophisticated methods for dropping data (Section \ref{seq:approx:posterior}). We assume throughout a $\GP$ with kernel \eqref{eqn5}.

\subsection{Naive sequential sampling}
\label{seq:seq:sampling}

{Posterior inference on $\f$ and hyperparameters $(\kappa,\alpha)$ is generally difficult. The likelihood of $\y$ may depend on $\f$ in a nonlinear way. In such cases, marginalisation of $\f$ is intractable, and inference relies on approximating the posterior distribution} $p(\f,\kappa,\alpha|\y) = \frac{1}{Z}p(\y|\f,\alpha)p(\f|\kappa)p_h(\kappa,\alpha)$
where $p_h(\kappa,\alpha)$ is a prior on $(\kappa,\alpha)$  and $Z$ is a normalising constant.

Efficient MCMC methods for sampling from the posterior for known or fixed $(\kappa,\alpha)$ are proposed in \citet{murray2010elliptical}. The hyperparameters may be inferred using techniques proposed in \citet{murray2010slice}. We will leverage these methods---ESS and  SDSS---but  consider a situation where sampling the full $\f$ is better avoided; either because of computational complexity or because the data arrives in a sequential manner. 

To this end, assume we have observed data $\y_1$; we generate an initial MCMC sample without making any approximations, using standard techniques as detailed at the end of Section \ref{seqAl}. Denote the initial MCMC sample of size $M$ by $\calS_1 =\{(\f_1^{(1)},\kappa^{(1)},\alpha^{(1)}),\dots, (\f_1^{(M)},\kappa^{(M)},\alpha^{(M)})\}$,
where $(\f_1^{(i)},\kappa^{(i)},\alpha^{(i)})$ is the $i$\textsuperscript{th} generated state of the Markov chain targeting the posterior $p(\f_1,\kappa,\alpha|\y_1)$.  

For the subsequent observation $\y_2$, a standard approach is to target the  joint posterior
\begin{equation}\label{eqn4}
p(\f_{1:2},\kappa,\alpha|\y_{1:2}) = \frac{1}{Z} p(\y_2|\f_2,\alpha)p(\f_2|\f_1,\kappa)p(\f_1,\kappa,\alpha|\y_1)
\end{equation}
{by blocked Gibbs sampling:} When iterating over $i\in\{1,\dots,M\}$, the state  $\f_2^{(i)}$ is first generated conditioned on $(\f_1^{(i)},\kappa^{(i)},\alpha^{(i)})$ by targeting its conditional distribution
\begin{equation} \label{eq:f:conditional}
p(\f_2|\f_1^{(i)},\kappa^{(i)},\alpha^{(i)},\y_{1:2}) \propto p(\y_2|\f_2,\alpha^{(i)}) p(\f_2 | \f_1^{(i)}, \kappa^{(i)} ).
\end{equation}
Second, $(\kappa^{(i)},\alpha^{(i)})$ is updated conditioned on $(\f_2^{(i)},\f_1^{(i)})$ by targeting its corresponding conditional, which includes every factor of \eqref{eqn4}. {Finally, the  state $\f_1^{}$ should also be revisited in an update conditional  on $(\f_2^{(i)},\kappa^{(i)},\alpha^{(i)})$.} 

In proceeding to steps $t=3,\dots,T$ we continue  to accumulate and update the latent states,
while hyperparameter updates target the  posterior  over the growing sequence $\f_{1:t}$, 
\begin{equation}\label{eqPostProp}
 \propto p(\y_t|\f_t,\alpha) p(\f_t | \f_{1:t-1}, \kappa )p(\f_{1:t-1}, \kappa,\alpha|\y_{1:t-1}).
\end{equation}
The complexity at each $t$ of this procedure is $\Ordo(t^3 N^3)$, which quickly becomes prohibitive as $t$ grows.

\subsection{Sequentially approximating the posterior}
\label{seq:approx:posterior}

In sequential data settings where, for example, data is collected daily, our task is often sequential, as well. When this is the case, previous $\f_s$, $s<t$, are no longer of interest; computation is saved by sampling only the most recent 
$\f_{t}$ from 
\begin{equation}\label{eqNoDrop}
\propto  p(\y_t|\f_t,\alpha) p(\f_{t} | \f_{1:t-1},\kappa),
\end{equation}
{i.e., ignoring updating previous $\f_s$ for $s<t$. The computational cost of the predictive prior in} \eqref{eqNoDrop} grows as $t^3$. If $\f_{t}$  has strongest prior  dependency with its most recent neighbours (as for isotropic kernels \eqref{eqn5} with respect to $t$), variables separated in time may be dropped in order to limit temporal dependency to the $\tau \geq 1$ most recent steps. The approximate predictive distribution is then
\begin{equation}\label{eqDrop}
p(\f_t|\f_{1:t-1},\kappa) \approx p(\f_t|\f_{t-\tau:t-1},\kappa)
\end{equation}
with a cost-cap of $\tau$. This naive form of  {data selection} has the purpose of limiting the size of $\f_{1:t-1}$ from growing with $t$ when evaluating the prior. $\tau$ can be set using domain knowledge or by tuning trade-off between computation and accuracy. For  sophisticated approaches to selecting \textit{which} latent variables to include in the predictive distribution, e.g., in the case of non-isotropic kernels, see \citet{osbornebayesian}. 

Sampling $\f_t$ is straightforward in the sequential procedure, as in \eqref{eq:f:conditional} for $t=2$; the computationally problematic term in \eqref{eqPostProp} is  the posterior from the previous time step. To circumvent the difficulties of parameter sampling,  we  approximate it by a factorised version
\begin{equation} \label{eq:approx:factor}
p(\f_{1:t-1},\kappa,\alpha|\y_{1:t-1}) \approx q_{t-1}(\f_{1:t-1})q_{t-1}(\kappa,\alpha)
\end{equation}
which is substituted { into} \eqref{eqPostProp}. 
The full conditional distribution for $\f_{t}$ is unchanged by the approximation, but the approximation yields simple updates for $(\kappa,\alpha)$ from the (umnormalized) target 
$p(\y_t|\f_t,\alpha)p(\f_{t}|\f_{1:t-1},\kappa)q_{t-1}(\kappa,\alpha)$.
Note that the approximate term $q_{t-1}(\f_{1:t-1})$ drops out of all sampling updates for $\f_t,\kappa,\alpha$, and so it does not need to be considered further. 

If $q_{t-1}(\kappa,\alpha)$ is a \textit{tractable}  approximation 
which can be updated in each $t$-step, the same basic sampling method can be applied sequentially by combining \eqref{eqDrop}-\eqref{eq:approx:factor}: (i) $(\f_{t},\kappa,\alpha) \sim p(\y_t|\f_t,\alpha)p(\f_{t}|\f_{t-\tau:t-1},\kappa)q_{t-1}(\kappa,\alpha)$, (ii) $q_{t-1}(\kappa,\alpha) \rightarrow q_{t}(\kappa,\alpha)$, (iii) $p(\f_{t}|\f_{t-\tau:t-1},\kappa) \rightarrow p(\f_{t+1}|\f_{t+1-\tau:t},\kappa)$,
using the priors $p_{h}(\kappa,\alpha)$ and $p(\f_1|\kappa)$ for the initial $t=1$. 
To make this scheme operational, we next consider a specific choice of  $q_{t-1}(\kappa,\alpha)$. 

\subsection{Sequential approximation and hyperparameter assumptions}
\label{sec:seq:ep}

Given the factorisation assumption \eqref{eq:approx:factor}, a number of methods for specifying the approximation $q_{t-1}(\kappa,\alpha)$ are possible. We propose an approach that seems natural given the problem constraints, based on ideas from expectation propagation \cite{minka2001expectation}. 

Consider minimizing the Kullback--Leibler divergence ${\KL(p(\f_{1:t},\kappa,\alpha | \y_{1:t}) || q_t(\kappa,\alpha))}$. The optimal solution is $  \hat{q}_t(\kappa,\alpha) = \int p(\f_{1:t},\kappa,\alpha | \y_{1:t}) d\f_{1:t} = p(\kappa,\alpha | \y_{1:t})$, which is  intractable in general. One tractable solution arises when $(\kappa,\alpha)$ are specified as deterministic transformations of a Gaussian vector $\z$ (an example is given below). 
Assuming a Gaussian approximating family, 
the optimal $\hat{q}_t(\z)$ is given by $\mathcal{N}(\m_{\z,t},\K_{\z,t})$ with  
\begin{align} \label{eq:moment:match}
  \m_{\z,t} = \bbE_{p(\z|\y_{1:t})}[\z] \quad \text{and} \quad 
  \K_{\z,t} = \bbE_{p(\z|\y_{1:t})}[\z \z']
\end{align}
i.e., by moment matching. At each step $t$, the sample $(\z_t^{(1)},\dotsc,\z_t^{(M)})$ can be used to re-estimate the moments \eqref{eq:moment:match}.
The approximation thus breaks the full temporal dependence into a sequential dependence similar to a Markov decomposition, and focuses computational resources via MCMC on parts of the model with more complex structure.

As an example specification for $\kappa$ and $\alpha$, we consider a scaled sigmoid Gaussian (SSG)
\begin{equation}\label{eqn7}
\kappa =\kappa_{\text{min}} + \frac{\kappa_{\text{max}}-\kappa_{\text{min}}}{1+\exp(-\z)},\quad\z\sim\mathcal{N}(\m_{\z},\K_{\z})\;,
\end{equation}
and similarly for $\alpha$. We make this choice because: (i)
it is convenient  for specifying the range of each parameter, $\kappa_i\in(\kappa_{\text{min},i},\kappa_{\text{max},i})$, as well as joint  dependency and distributional shape with $\m_{\z},\,\K_{\z}$; 
and (ii) as the parameter vector is effectively a transformed Gaussian, we can leverage ESS when updating hyperparameters in the sampling scheme. 
For a non-informative prior one may  set $\m_{\z}=0$ and  $\K_{\z}=1.5^2\mathbb{I}$ while
domain knowledge can be used for setting $\kappa_{\text{max}}$ and $\alpha_{\text{max}}$. We can then take our posterior representation $q_{{t}}(\kappa,\alpha)$ to also be a SSG with $\m_\z$ and $\K_\z$ updated from the previous hyperparameter sample, i.e., moments estimated from the posterior sample $\{(\kappa^{(1)},\alpha^{(1)}),\dots,(\kappa^{(M)},\alpha^{(M)})\}$, generated at  $t$.

\subsection{Algorithm}\label{seqAl}
The workhorse of our algorithm is the elliptical slice sampler. For a given $t$, we  use it to sample from the target approximative posterior,
\begin{equation}\label{eqn9}
\pi(\f_t,\kappa,\alpha|\y_t) = \frac{1}{Z} p(\y_t|\f_t,\alpha)p(\f_t|\f_{t-\tau:t-1},\kappa)q_{t-1}(\kappa,\alpha)
\end{equation}
 in three steps (see Algorithm \ref{alg2}):
 
 \textbf{Update 1: $\f_t$ for fixed $(\kappa,\,\alpha,\f_{t-\tau:t-1})$ with ESS.} The  covariance matrix $\K_{t,\kappa}$ (square-root decomposition $\K_{t,\kappa} = \bL_{t,\kappa}\bL^{\top}_{t,\kappa}$) and mean $\m_{t,\kappa}$ of the conditional prior  $\f_t|\f_{t-\tau:t-1}\sim\mathcal{N}(\m_{t,\kappa},\K_{t,\kappa})$ must be computed only once, using standard predictive equations for Gaussian processes \cite{rasmussen2006gaussian}.
 Since $\kappa$ is fixed, $\K_{t,\kappa}$ and $\m_{t,\kappa}$ stay unchanged when applying ESS so it is sensible to repeat this operation and update $\f_t$ several times. 
 
  \textbf{Update 2: Covariance parameters $\kappa$ for fixed $\alpha$.} We represent the conditional prior of $\f_t$ as a transformation of a spherical Gaussian,
 \begin{equation}\label{eqnHejDu}
\f_t = \bL_{t,\kappa}\bnu + \m_{t,\kappa},\quad\bnu\sim\mathcal{N}(0,\mathbb{I})
\end{equation}
and sample $\kappa$ for fixed $\bnu$ with respect to $p(\y_t|\f_t=\bL_{t,\kappa}\bnu + \m_{t,\kappa},\alpha)q_{t-1}(\kappa,\alpha)$. Note that this  will also update $\f_t$ as a by-product. Due to the posterior representation  of $\kappa$ with a SSG \eqref{eqn7}, we perform this update in $\z$-space, again with ESS targeting $p(\y_t|\f_t(\z),\alpha)\mathcal{N}(\z;\m_\z,\K_\z)$ where $\f_t(\z)=\bL_{t,\kappa=\text{ssg}(\z)}\bnu + \m_{t,\kappa=\text{ssg}(\z)}$. 

 \textbf{Update 3: Likelihood parameters $\alpha$ with fixed $(\kappa,\,\f_t)$.} 
As with the update for $\kappa$, this update is done in the underlying $\z$-space, with ESS targeting $p(\y_t|\f_t,\alpha=\text{ssg}(\z))\mathcal{N}(\z;\m_\z,\K_\z)$. Note that this update does not involve the conditional prior of $\f_t$. Thus, it can be relatively cheap since it only requires likelihood evaluations.  

In order to enhance exploration, before stepping to the next $t$, it is sensible to perform  additional updates of $\f_t$, given the updated parameters.  This may also be done in an intermediate update between Update 1 and  Update 2. Algorithm \ref{alg2} summarises all steps. To sample the full $\f_{1:T}$ and hyperparameters, we apply the above three updates for $i=1,\dots,M$ in an inner loop to obtain the sample $\mathcal{S}_t$ given $\mathcal{S}_{t-1}$, and sequentially for $t=2,\dots,T$ in an outer loop.

\paragraph{Initial sample} When generating $\mathcal{S}_1$ it is worth investing in a Markov chain with a  large number of states 
to ensure a good representation  of the initial posterior. Generated variables and parameters are used for conditioning $\f_2$ and for re-estimating the posterior of $(\kappa,\alpha)$. Updates of the subsequent $\mathcal{S}_2$ are therefore more efficient if the transient phase of the first chain has been discarded as burn-in. One also benefits from thinning the collected samples, both for reducing the dependence between samples and to control the size  of $M$ for all subsequent samples.

The strength of the  prior over $\f_1$ is often weak relative to the likelihood; updating covariance parameters with a fixed-$\bnu$ representation (\ref{eqnHejDu}) therefore is likely to mix poorly. This issue is discussed in \citet{murray2010slice} who propose 
the more efficient SDSS. Thus, it is advantageous to use their method for sampling parameters at the first $t=1$; especially since we can adopt an ESS update (in $\z$-space) within SDSS under our hyperprior/posterior assumptions. For subsequent $t$, however, we expect $\f_t$ to have strong ties with $\f_{t-\tau:t-1}$. Therefore, we can indeed expect to have an informative (conditional) prior, such that a fixed-$\bnu$ update is likely to be efficient. In doubt, one may always apply SDSS for every time step.

\begin{algorithm}
   \caption{Sequential sampling}
   \label{alg2}
   {\bfseries Input:} Initial sample $\mathcal{S}_1$ of size $M$.
    {\bfseries Output:} Samples $\mathcal{S}_2,\dots,\mathcal{S}_T$.
\begin{algorithmic}[1]
\FOR{$t=2$ to $T$}
\STATE Update: $\m_\z = \text{mean}(\z_{t-1})$, $\K_\z = \text{cov}(\z_{t-1},\z_{t-1})$ with $\z_{t-1} = \text{ssg}^{-1}((\kappa^{(1:M)};\alpha^{(1:M)}))$ from $\mathcal{S}_{t-1}$; likelihood function $p(\y_t|\cdot)$  based on new data $\y_t$.
   \FOR{$i=1$ to $M$}
   \STATE Initiate: $\f_{t-1}^{(i)}$ from $\mathcal{S}_{t-1}$; $\kappa,\alpha$ from most recent updates. 
   \STATE Initial draw: $\f_{t}^{(i)}\sim\mathcal{N}(\m_{t,\kappa},\K_{t,\kappa})$ with $\m_{t,\kappa}$, $\K_{t,\kappa}$ calculated from $\kappa$ and $\f_{t-1}^{(i)}$.
   \STATE Update 1: $\f_t^{(i)}\sim \text{ESS}(\f_t^{(i)};\m_{t,\kappa},\K_{t,\kappa})$ $\times$1--10.
   \STATE Update 2: $\kappa^{(i)},\f_t^{(i)}\sim\text{ESS}(\kappa;\m_\z,\K_\z)$ with fixed $\bnu$. 
   \STATE Repeat update 1.
   \STATE Update 3: $\alpha^{(i)}\sim\text{ESS}(\alpha;\m_\z,\K_\z)$.
      \STATE Repeat update 1.
   \ENDFOR
   \STATE	Save: $\mathcal{S}_t = \{\f_t^{(i)},\kappa^{(i)},\alpha^{(i)}\}_{i=1}^M$
\ENDFOR
\end{algorithmic}
\end{algorithm}

\section{Conceptual analysis} \label{sec:concepts}

Detailed theoretical analysis of the sequential approximation is beyond the scope of this paper. Rigorous theoretical theoretical treatment of MCMC sampling from an approximate posterior can be found in, e.g., \citet{Pillai:2014aa,Johndrow:2015aa,Johndrow:2017aa}. Here we consider our method from a high-level conceptual perspective. 

The proposed method introduces an approximation of the true posterior at each time step. By limiting the temporal dependence between the observations in the current time step and those in previous time steps, the method induces a smaller effective sample size for performing inference at the current time step. As a result, the variance of the approximate posterior generally will be larger than that of the true posterior. Such an effect is seen in the experiments, particularly in Figure \ref{fig9}. In practice, restricting the influence of earlier time steps may have beneficial effects, particularly if there is model mis-specification, e.g., if the data are not stationary. Related ideas are explored in broader context in \citet{Jacob:2017aa}. As we show below, if the ratio of the approximate posterior variance to the true posterior variance grows sub-quadratically in the number of time steps, $t$, then the sequential approximation may reduce overall error.

\paragraph{Bias-variance trade-off} The sequential algorithm proposed in Section \ref{sec:lgpm} introduces bias: At each $t$, the method generates exact samples from an approximate posterior distribution. Conversely, estimates made with those samples will exhibit smaller variance than an unbiased MCMC method due to the larger number of samples generated by the sequential approximation per unit of computation time. In particular, let $\varphi_t$ be some function of $\f_{t},\kappa,\alpha$, and suppose we wish to compute $I^{\varphi_t} = \bbE_{p(\f_{t},\kappa,\alpha \mid \y_{1:t})}[\varphi]$. Let ${\IhatSGP(M) := \frac{1}{M}\sum_{i=1}^M \varphi_t(\f_t^{i},\kappa^{i},\alpha^{i})}$ denote the estimate of $I^{\varphi_t}$ based on $M$ independent samples from the sequential GP approximation, and likewise for $\IhatMC(M)$ based on an unbiased MCMC scheme. The mean square error (MSE) of $\IhatSGP(M)$ is
\begin{align*}
	\mse(\IhatSGP(M)) & = \bbE[(\IhatSGP(M) - \I)^2] \\
	  & = \Var[\IhatSGP(M)] + \bias^2 \;,
\end{align*}
where $\bias$ is the bias of the sequential approximation after $t$ time steps. For an unbiased MCMC method, $\mse(\IhatMC(M)) = \Var[\IhatMC(M)]$. Therefore, assuming that $\Var_{p(\f_{t},\kappa,\alpha \mid \y_{1:t})}[\varphi_t] = \sigma_t^2 < \infty$ for the true posterior and likewise $\sigma_{\text{SGP},t}^2 < \infty$ for the sequential approximation,
\begin{align*}
 \Delta & :=  \mse(\IhatMC(M)) - \mse(\IhatSGP(M')) \\
 	& = \frac{\sigma_t^2}{M} - \frac{\sigma_{\text{SGP},t}^2}{M'} - \bias^2 \;.
\end{align*}
For a fixed amount of computation time, $t_c$, $\Delta$ may be approximated using the complexity of the sampling algorithms (with $C$ and $C'$ capturing the hidden constants for MCMC and sequential sampling, respectively),
\begin{align*}
	\Delta \approx \tilde{\Delta} := \sigma_t^2 \frac{C t^3 N^3}{t_c} - \sigma_{\text{SGP},t}^2\frac{C' t\tau^3 N^3}{t_c} - \bias^2 \;.
\end{align*}
As discussed above, the approximate posterior will in general have higher variance than the true posterior; denote the ratio of the two as $R_t := \sigma_{\text{SGP},t}^2 / \sigma_t^2 \geq 1$. 
Therefore, the sequential approximation yields lower error than unbiased MCMC if $\tilde{\Delta} > 0$; equivalently,
\begin{align} \label{eq:bias:variance}
 \bias^2 < \frac{\sigma_t^2 t N^3}{t_c}(  C t^2 - C' R_t \tau^3 ) \;.
\end{align}
In order for the r.h.s. to be positive, $R_t < \frac{Ct^2}{C'\tau^3}$. 
Experiments indicate that $R_t$ is relatively stable even for small $t$ (see Figure \ref{fig9}); assuming positivity for large enough $t$ indicates that the sequential approximation will reduce MSE if the bias grows at most as $\sqrt{t}$. The experiments in Section~\ref{sec:experiments} provide evidence that the approximation bias does not grow quickly for the cases considered there.

\section{Experiments} \label{sec:experiments}

To demonstrate our approach, we apply sequential sampling to a $\GP$ regression model on synthetic data in the next section. For brevity, we omit further standard applications, such as $\GP$ classification and Cox-process inference. Instead we concentrate on a more complicated financial model with  a nonlinear likelihood from an option pricing problem. 

\subsection{Gaussian Process regression {for 3D inputs}}\label{seqHej}
We consider the Gaussian regression problem from \citet{murray2010elliptical} as our starting point. For each $t$, data  $\y_t$ 
are noisy observations of latent values $\f_t$ 
taken at input locations $\x_t$. We set up the data to have $N=200$ observations for each $t$, with inputs $\x_t$ drawn uniformly over a unit square. Latent values are generated from a $\GP$ prior with squared-exponential kernel of parameters $\kappa=(l_{x_1},l_{x_2},l_{t},\sigma_f)$,
$$k(t,x;t',x') = \sigma_f^2 \exp\left(-\sum_{d=1}^2 \frac{(x_{(d)}-x'_{(d)})^2}{2l_{x_d}^{2}}-\frac{(t-t')^2}{2l_t^2}\right).$$
We let $\sigma_f=1$ and draw length-scales uniformly over $(0,\sqrt{10})$. We set $(\mu_f,\sigma_y)=(0.5,.3)$  for the  likelihood 
$ p(\y_t|\f_t,\alpha) = \mathcal{N}(\y_t|\f_t+\mu_f,\sigma_y^2\mathbb{I})$, and $\kappa_{\text{max}}=(\sqrt{10},\sqrt{10},\sqrt{10},2)$ and $\alpha_{\text{max}}=(1,1)$ 
for the hyperprior. 
Finally,  we generate the full data set with {$T=20$ steps of $t$} equally spaced over the unit interval.  Each of $\y_{1:T}$, $\f_{1:T}$, $\x_{1:T}$, thus has $T N=4000$ elements.

For $t=1$, we generate 6000 states by the algorithm outlined in Section \ref{seqAl} for the initial sample, with three $\f_1\sim \text{ESS}$ updates in the first update (running time $<4$ minutes on a 2.8GHz quad-core Intel i7 processor). Note that we target  the exact posterior and therefore $l_t$ can not be inferred as we see observed data for a singe $t$ only.   We discard the first 1000 states as burn-in and keep every $5^{\text{th}}$ state to obtain a thinned sample $\mathcal{S}_1$ of size $M=1000$. We then continue to generate samples $\mathcal{S}_2,\mathcal{S}_3,\dots,\mathcal{S}_{20}$ sequentially from their approximative posteriors: For each $t$, we 
generate $\mathcal{S}_t$ from  the target $\pi(\f_t,\kappa,\alpha|\y_t)$ in (\ref{eqn9}) based on the previous states of $\mathcal{S}_{t-1}$. We drop all but $\tau=1$ variables. In each iteration, we repeat $\f_t\sim\text{ESS}\times5$ in the first update, and another five repetitions between the second and third update of the parameters (total running time 35 minutes).

\begin{figure}
\centering
\includegraphics[width=\columnwidth,trim=0 10 0 50,clip]{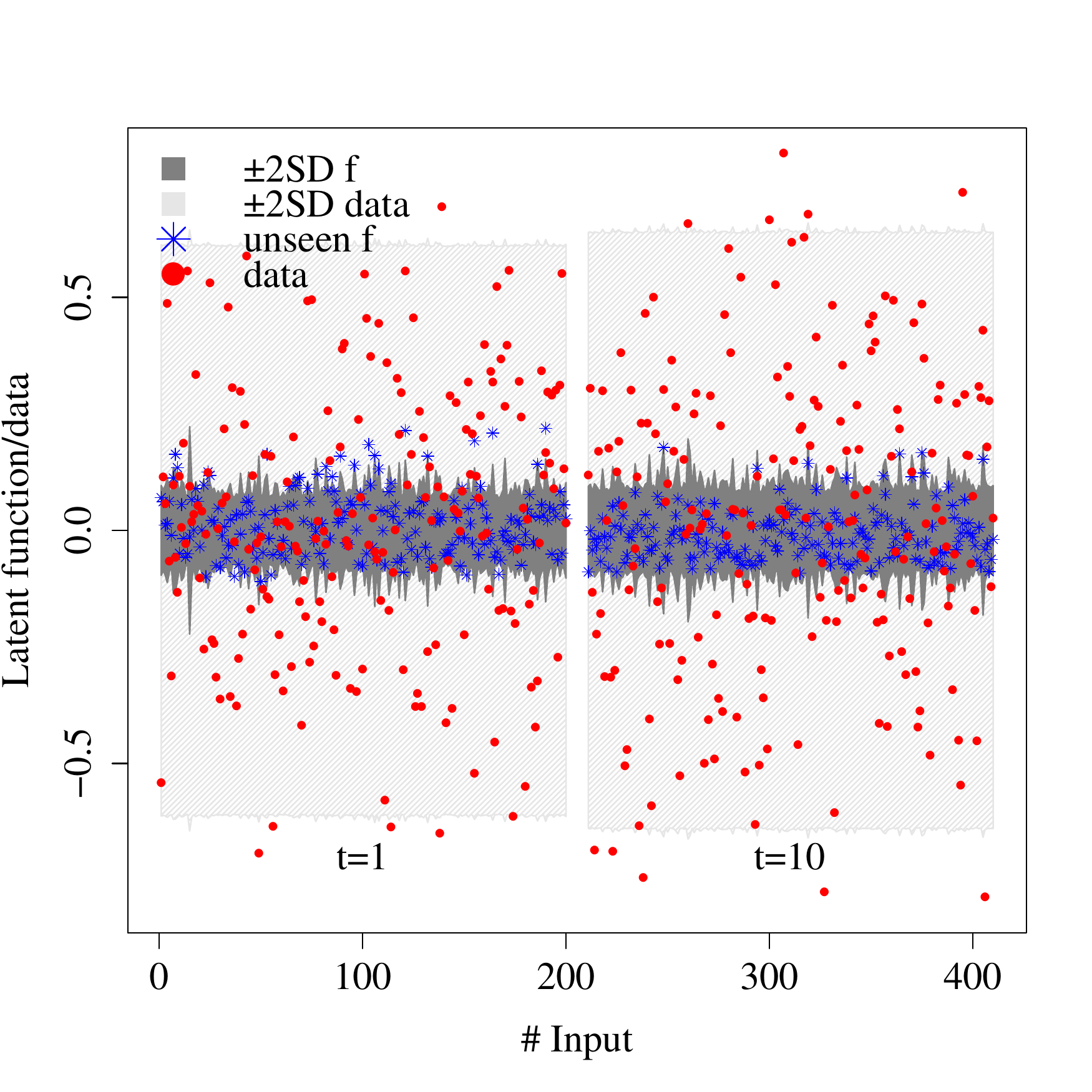}
\caption{ Inputs  1--200 show  $\mathcal{S}_1$ from the GP regression problem. The posterior over $\f_1$ is represented  
 by a $\pm2\text{SD}$ region in dark grey. Light grey is the $\pm2\text{SD}$ credible interval for corresponding  data. For clarity, the sample mean of $\f_1$ is subtracted from all { values}, such that the posterior is centered around zero.   Inputs 201--400 show corresponding results for  $\mathcal{S}_{20}$. }
\label{fig2}
\end{figure}

\textbf{Posterior representation}\, Input \#1--200 of  {Figure} \ref{fig2} represents the  posterior over the initial  latent variable $\f_1$  ($\mu_f$ included) from $\mathcal{S}_1$. The sample captures  the latent process to a good extent: 85\% of true values (blue stars) fall within the  $\pm2$SD region. A credible interval for $\y_1$ is also calculated from the  sample of $\f_1$ and $\sigma_y$ (light grey). Comparing to observed data (red dots), we see a good representation with a few data points lying outside the interval (8 out of 200). Sequential samples over $\f_t$ and credible intervals for $\y_t$ yield similar results for each $t$ as  seen for $\mathcal{S}_1$. Plots are not shown here for brevity, except for results from  $\mathcal{S}_{20}$  shown as \#201--400 in  Figure \ref{fig2}. The samples exhibit a good representation of the ground truth: all latent values fall inside the  $\pm2$SD posterior sample, and data observations fall inside their  $\pm2$SD credible interval (9 out of 200). 

\begin{figure}
\includegraphics[width=0.49\textwidth]{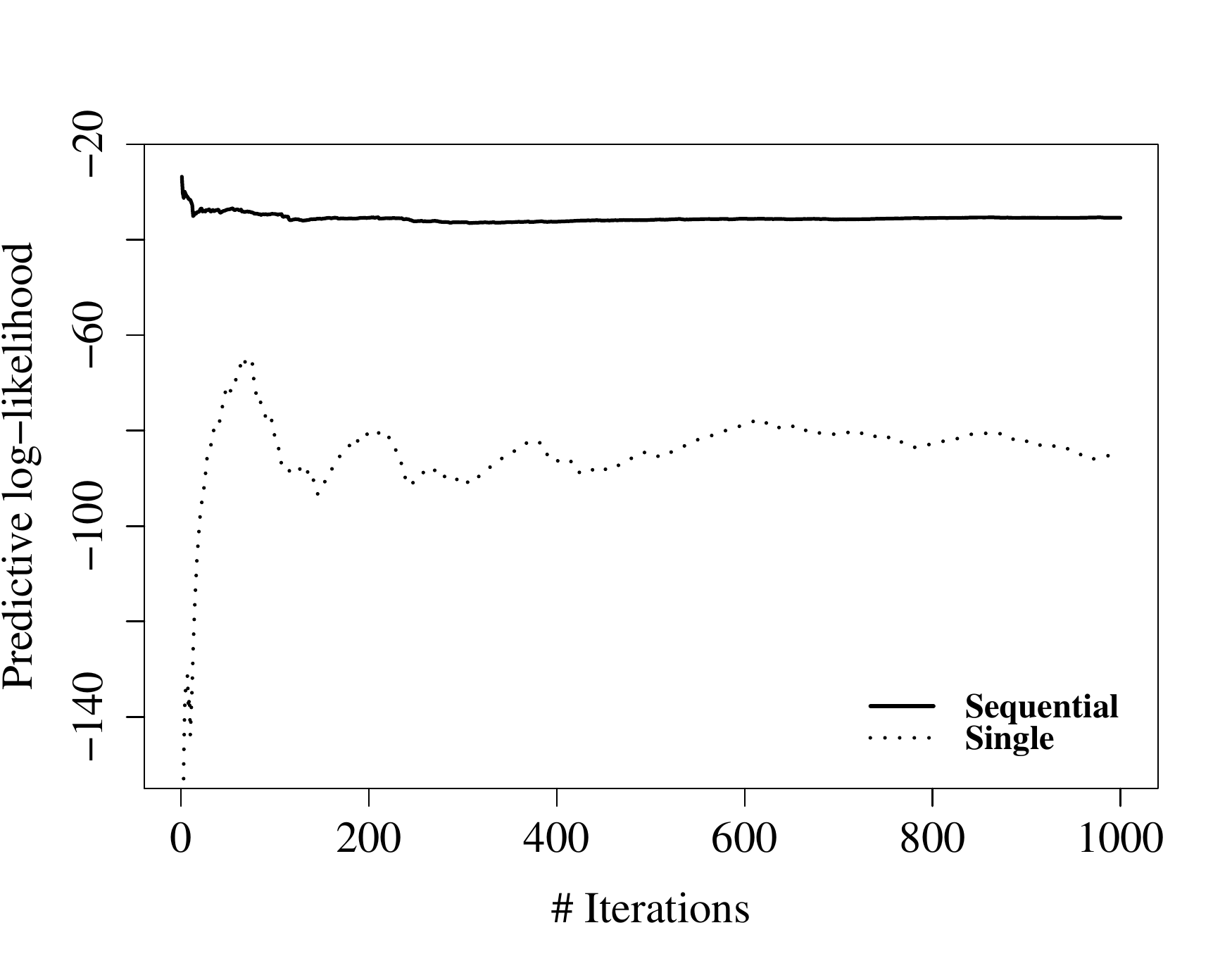}
\caption{$\GP$ regression. 
The predictive distribution $\f_{20}^{\star}|\f_{20}^{o(i)}$ is based on $\f_{20}^{o(i)}$ sampled with the sequential approximation (solid line), and from a single MCMC with no dependence on previous time steps for $t=20$ (dotted line), respectively. The higher predictive log-likelihood of the sequential approximation indicates that it captures statistical information from previous time steps.}
\label{fig4}
\end{figure}

\textbf{Posterior prediction}\, To further demonstrate the applicability of our approach, we perform a prediction experiment as follows. Given  samples up to and including $\mathcal{S}_{t-1}$, we divide the subsequent data into a training set $\y_t^o$  and a test set $\y_t^\star$ of equal size.\footnote{This is simply done by assigning all input points of $\x_t$ lying in the half plane $[0,0.5]\times[0,1]$ to the training set, and remaining points to the test set.} We  sample $\f_t^o$ from the  target (\ref{eqn9}) with likelihood given by $\y_t^o$. From each such sample $\f_t^{o(i)}$, we  compute the  predictive mean $\m_t^{\star(i)}$ and covariance $\K_t^{\star(i)}$ of the test variables $\f_t^\star|\f_t^{o(i)}$ 
and represent their predictive power by the predictive likelihood of $\y_t^{\star}$, based on $\m_t^{\star(i)}$ and $\K_t^{\star(i)}$. Denoting the corresponding log-likelihood value $ll^{\star(i)}$, we then look at how the accumulative mean of $ll^{\star(1:i)}$ varies with sampling iteration $i$. This gives an idea of how predictive power changes as an increasing number posterior samples of $\f_t^{o}$ are used; Figure \ref{fig4} shows the result for $t=20$.

For comparison, we repeat the experiment but with no previous information from $\mathcal{S}_{t-1}$. That is, we use the ``initial'' sampling scheme based only on data $\y_t^o$, breaking all dependence with earlier time steps. The resulting predictive likelihood of $\y_t^{\star}$---shown in Figure \ref{fig4} for $t=20$---is now weaker, indicating that the sequential approximation is capturing information from previous time steps. 

\begin{figure*}
\centerline{
\includegraphics[width=0.9\textwidth,trim=0 10 0 50,clip]{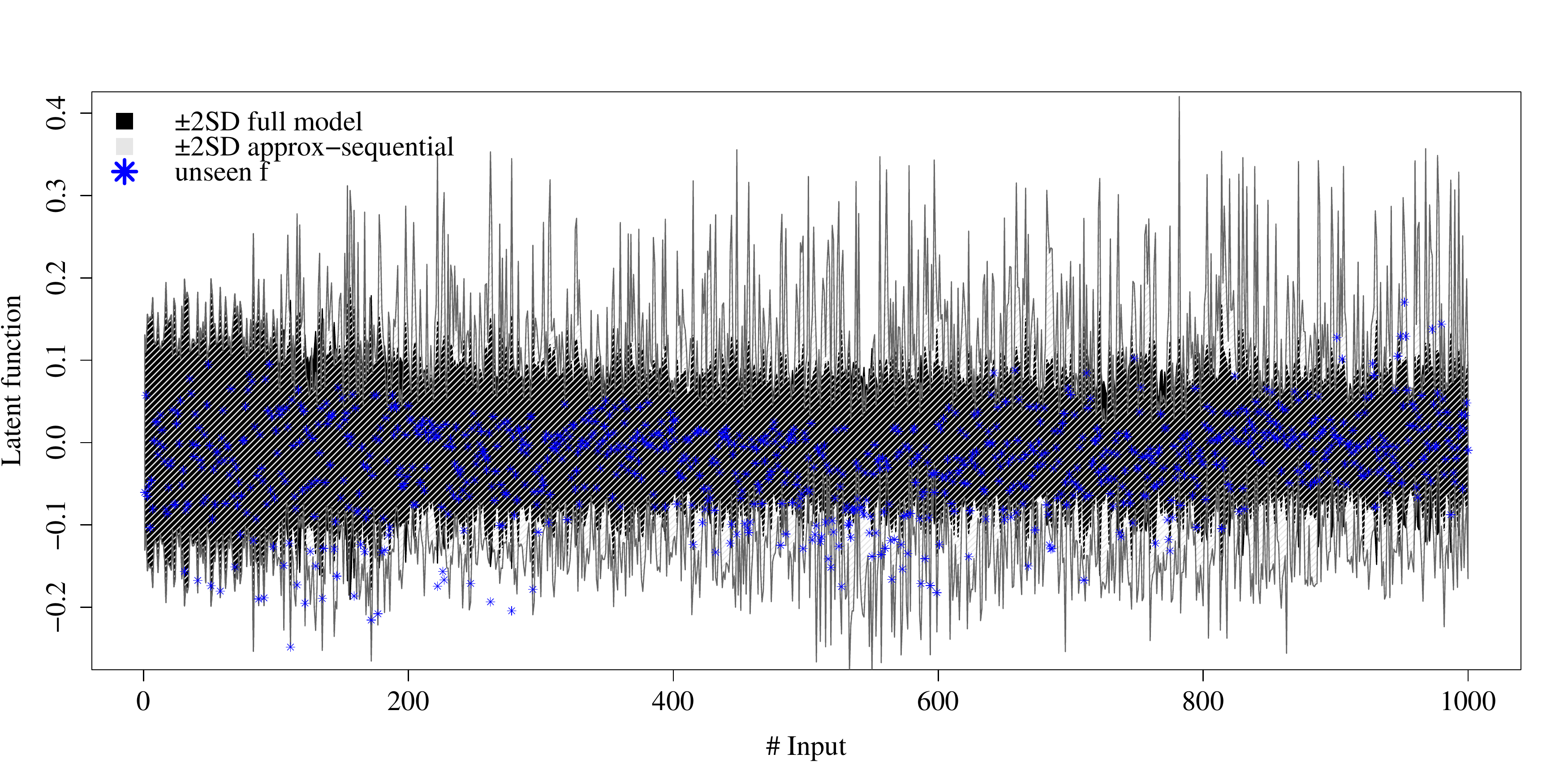}
}
\caption{Comparison of approximative-sequential sampling vs. sampling the full model.  }
\label{fig9}
\end{figure*}

\textbf{Comparison with full sampling}\, For comparison, we use the full, non-approximate MCMC method on a $\GP$ regression model with $N=100$ and $T=10$. 
We sample the full model---non-sequentially, without posterior approximations---with the \textbf{initial sample} method. We generate 6000 MCMC samples of $\f_{1:t}$ for $t=1,\dots,10$; computation time is 1 minute for $\f_1$, up to 11 hours for $\f_{10}$. Similarly, we generate 6000 samples with sequential sampling, which takes 3 minutes. The sub-sampled result is illustrated in Figure \ref{fig9}. In qualitative terms, the main difference is that the overall variance decreases in $t$ for the full sampling method because it uses the full data set $\y_{1:t}$, 
while sequential sampling has a constant, higher sample variance due to the approximations, which limit temporal dependence.  
Note also that the sequential posterior does not deteriorate with $t$, indicating that  sequential approximation does not introduce a bias which \textit{accumulates}. Similarly it does not drift, such that the bias and variance are \textit{stable} across the sequence.

\subsection{Option pricing problem}
As a second application, we consider inferring a latent positive function $\sigma(T,K,t)$ of an option pricing model \cite{dupire1994pricing}. For fixed $t$, the model is represented by a mapping $\sigma(\cdot,t)\mapsto C(\cdot)$ where the price function $C(T,K)$ solves the PDE
\begin{align}\label{dupire}
\frac{\partial C}{\partial T}+rK\frac{\partial C}{\partial K} - \frac{K^2\sigma^2(T,K)}{2}\frac{\partial^2C}{\partial K^2} = 0
\end{align}
with boundary condition $C(0,K)=(S_t-K)^+$. The function $C(T,K)$ yields the time-$t$ price of a call option with maturity $T$ and strike $K$ on an underlying asset with current price $S_t$. 

The construction and calibration of $\sigma$ from a set of observed market prices at a single date $t$  is a  problem commonly encountered in quantitative finance---see for instance the discussion by \citet{luo2010local}. We consider it here as an example of a challenging likelihood and inference problem naturally placed in a sequential context. Given $\bsigma_t=\sigma(\x_t)$, where $\x_t$ is the set of $(T,K)$-inputs observed at $t$, we take observed call prices  $\c_t$ to be generated with Gaussian noise
\begin{equation*}
p(\c_t|\bsigma_t) = \mathcal{N}(\c_t;C(\x_t;\bsigma_t),\sigma^2_c\mathbb{I}).
\end{equation*}
Here, $C(\x_t;\bsigma_t)$ are model prices for each strike-maturity in $\x_t$, calculated with $\bsigma_t$. Further, we place a zero-mean $\GP$ prior on $\f=\{ \f_t\}_{t=1}^T$ and use the ``softplus'' function $\zeta(f)=\log(1+\exp(f))$  to impose positiveness; $\bsigma_t =\zeta(\f_t+\mu_f)$. 
In effect, the likelihood factorises over time components $p(\c_t|\zeta(\f_t+\mu_f))$. However, it does not factorise over the components of $\f_t$,  it is highly nonlinear (thus non-Gaussian) and intractable---it can not even be evaluated with a closed-form expression.\footnote{ Since no closed-form solution of (\ref{dupire}) is known for a general function $\sigma$, we follow standard procedure and use a numerical Crank--Nicolson solver \citep[see, e.g.,][]{hirsa2012computational}.} 

We generate data with $T=12$ equidistant time steps and $N=75$ observations for each $t$ from $\x_t$ placed at a grid of 15 strikes $\times$ 5 maturities. Each $C(\x_t;\bsigma_t)$ is computed with $\bsigma_t$ from a draw $\f_{1:T}$  of  a $\GP$ with squared-exponential kerned.  We use $(l_T,l_K,l_t,\sigma_f)=(0.5,0.3,0.5,0.75)$ and $(\mu_f,\sigma_c)=(-1.5,0.05)$ for the likelihood. For the prior, $\kappa_{\text{max}}=(1,1,1,1)$ and $\alpha_{\text{max}}=(0.5,0.5)$. We set an interest rate $r=0$ while the underlying price $(S_t)_{t=1}^{T}$ is simulated from a geometric Brownian motion with $(S_1,\mu,\sigma)=(1000,0.04,0.2)$.

Sequential sampling from the posterior is carried out in the same manner as for the regression problem of Section \ref{seqHej}. We generate 20,000 states for $t=1$ (running time 18min), discard 10\% as burn-in and subsample to obtain $\mathcal{S}_1$ of size $1000$ before continuing with $\mathcal{S}_2,\dots,\mathcal{S}_{12}$ (total time 24min).

\begin{figure}[t]
\vskip 0.2in
\begin{center}
\centerline{\includegraphics[width=\columnwidth,trim=0 10 0 50,clip]{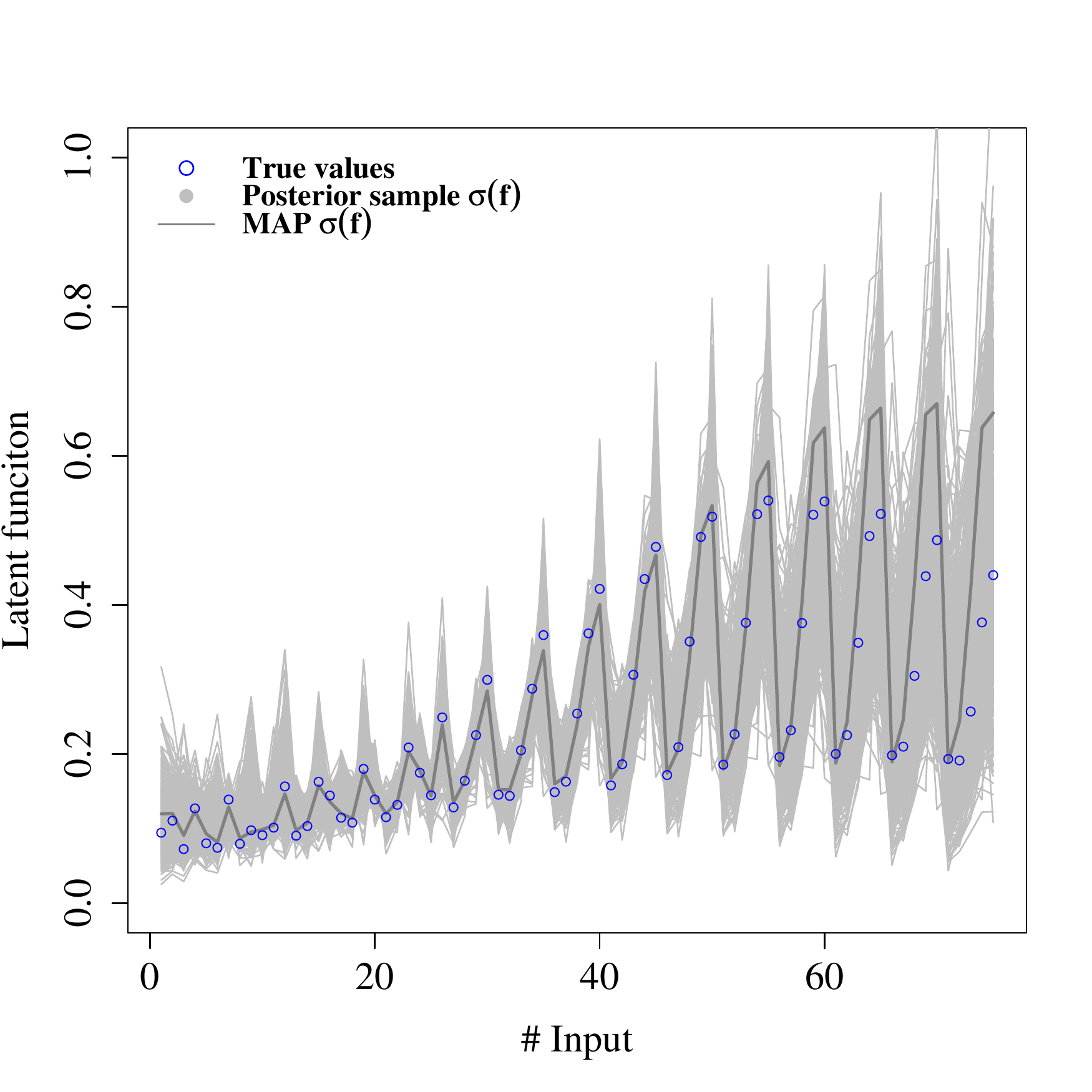}}
\caption{The inferred latent function $\bsigma_{12}=\zeta(\f_{12}+\mu_f)$ from $\mathcal{S}_{12}$ for the option pricing problem.  The posterior sample is shown in grey, its MAP estimate with a solid grey line and true latent values with blue circles. }
\label{fig5}
\end{center}
\vskip -0.2in
\end{figure}

\begin{figure}[t]
\vskip 0.2in
\begin{center}
\centerline{\includegraphics[width=\columnwidth,trim=0 10 0 50,clip]{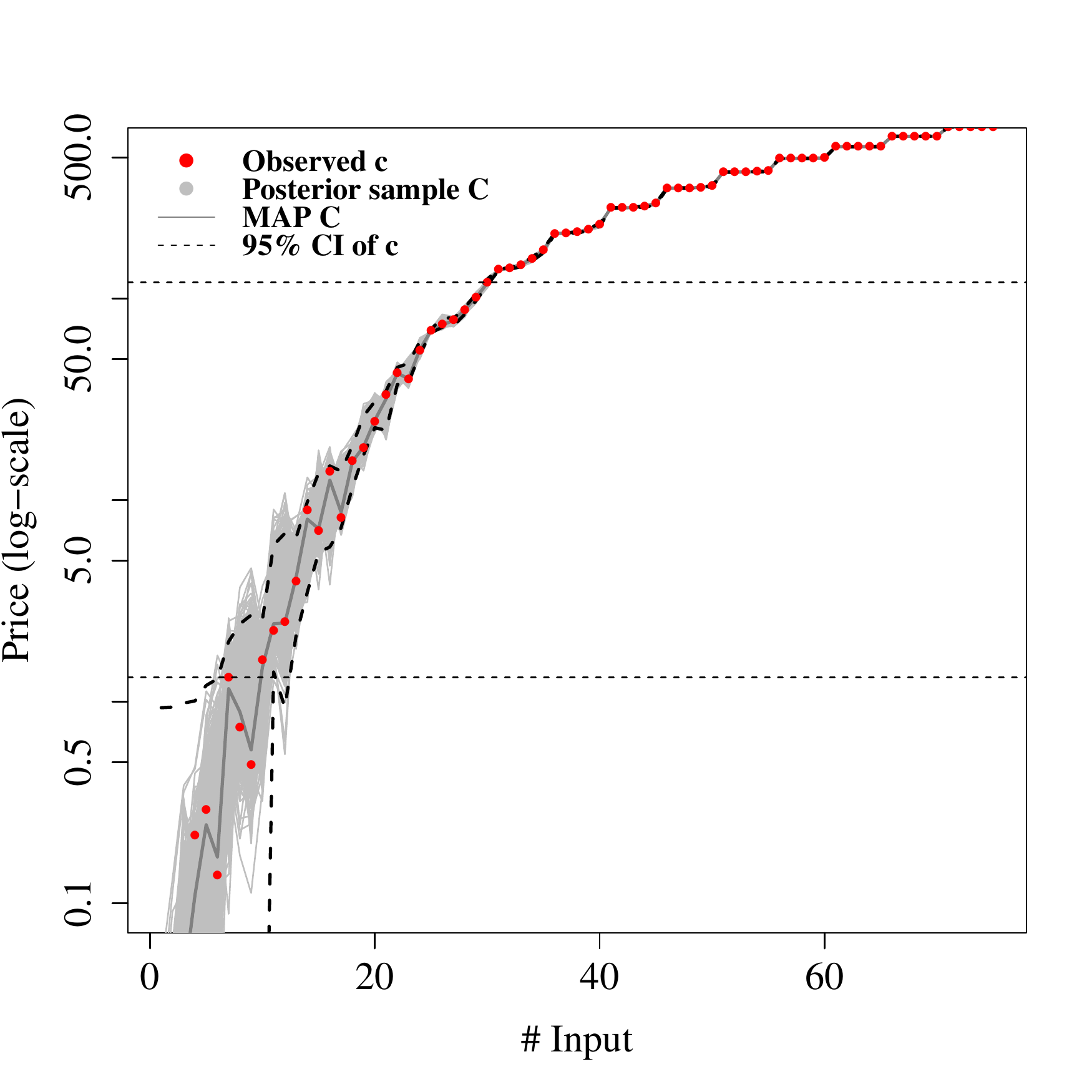}}
\caption{Option pricing problem; $t=12$ sample. The estimated posterior of the model price is shown in logarithmic scale (grey area), corresponding to the latent function shown in Figure \ref{fig5}. Prices are plotted in groups of five maturities with common strike (strike-price in descending order with \# Input). Thick dashed lines represents a posterior 95\% credible interval for observable prices (model price + noise) while red dots shows observed values used for inference. The two horizontal lines indicates which options are close to at-the-money.}
\label{fig6}
\end{center}
\vskip -0.2in
\end{figure}

\textbf{Posterior representation}\, For the last sequential step, 
Figure \ref{fig5} shows the posterior sample of latent variables $\bsigma_{12}=\zeta(\f_{12}+\mu_f)$. The posterior sample  covers true latent values to a satisfactory extent. We also see that the MAP estimate of $\bsigma_{12}$ is close to true values.   More interesting is that the \textit{uncertainty} in the posterior clearly varies over the input space. This is a consequence of the non-linearity of the likelihood, since the sensitivity of $C$ with respect to $\sigma$ varies with maturity and strike (c.f. variability over latent variables in Figure \ref{fig2}). In Figures \ref{fig5} and \ref{fig6}, inputs are ordered in groups of five with common strike. Taking one such group from the right hand side of Figure \ref{fig5}, where \emph{strikes are low}, the variability over $\bsigma$ is large. The reason is that a low strikes gives a call option deep \textit{`in-the-money'}, such that its pay-off is relatively certain, and thereby its price insensitive to $\bsigma$. In effect, the likelihood is relatively uninformative about $\bsigma$ in low-strike regions.


The corresponding posterior sample over option prices $\c_{12}$ is shown in right 
Figure \ref{fig6} in log-scale. The non-linearity effect of the pricing function is  clearly manifested: even if there is large uncertainty about $\bsigma$ for {low} strikes 
(Figure \ref{fig5}, \# Input $>$ 40) there is  little  uncertainty over corresponding prices 
(Figure \ref{fig6}, \# Input $>$ 40, albeit the log-scale). More interesting are the options around \textit{at-the-money}, with prices between the two horizontal  lines (inputs $\sim$ 10--30). These options are the most actively traded, and hence a good fit of model to market prices is desirable. Taking the MAP estimate (the sample surface $\bsigma_{12}^{(i)}$ which achieves largest posterior likelihood) this is indeed the case: Observed data falls close to MAP prices in Figure \ref{fig6}. Further, the full posterior sample demonstrates how \textit{parameter uncertainty}---inherent in a model when estimated from data---is distributed across strikes and maturities. This representation of uncertainty is important, not the least as it should be taken into account when the model is used for prediction. Finally, we note that the results for $t\in\{1,\dots,11\}$ are very similar in quality to those discussed in the above, but not shown for brevity: the efficiency of our sequential sampling procedure is consistent across $t$.

\section{Conclusion and discussion}

We have proposed a computationally efficient sampling strategy that applies to Bayesian inference for $\GP$-based latent variable models  with sequentially increasing data. We proposed a practical approach based on a strategic approximation that: (i) breaks the joint posterior from the previously sampled step into its marginals over latent variables and parameters; (ii) represents this parameter marginal with a transformed Gaussian to enable it being updated from its the recent sample; and (iii) drops variable in the conditional prior over latent variables. We demonstrated its benefits for a standard $\GP$ regression model on synthetic data  of size that would be impractical with standard sampling, and for a complicated option-pricing model with highly nonlinear likelihood. Both examples showed strong performance of our method, with good posterior representation of the ground truth. For the regression problem, we also showed that it is competitive with full sampling, at a massive reduction in computation time.

Our sampling  scheme  will not produce outcomes from the true posterior distribution as it targets the approximation (\ref{eqn9}). This is the price we have to pay for computational efficiency. The approximation will perform best when the data are highly informative: $p(\f_{1:t} | \y_{1:t}, \kappa,\alpha)\approx p(\f_{1:t} | \y_{1:t})$ and when there is an isotropic dependency structure over time. In our examples, this is the case. The savings in computation can be substantial: If the conditional prior of $\f_t\in\mathbb{R}^N$ is capped to $\tau$ previous labels, we have a reduced  overall complexity  from $\mathcal{O}(T^3N^3)$ to $\mathcal{O}(T\tau^3N^3)$ for $T$ sequential updates.

\clearpage

\section*{Acknowledgments}
The research leading to these results has received funding from the Oxford-Man Institute of Quantitative Finance (MT and SR), and
European Research Council under the European Union's Seventh Framework
Programme (FP7/2007-2013) ERC grant agreement no. 617071 (BBR).

\bibliography{../example_paper}
\bibliographystyle{icml2018}

\end{document}